\documentclass[letterpaper, 10 pt, conference]{ieeeconf}  % Comment this line out if you need a4paper

\IEEEoverridecommandlockouts                              % This command is only needed if 
\usepackage[utf8]{inputenc}
\usepackage{xspace}
\usepackage{comment}
\usepackage{graphicx}
\usepackage{xcolor}
\definecolor{gg}{RGB}{0, 155, 85}
\definecolor{primarycolor}{RGB}{33,49,77}   % dark blue
\definecolor{angrycolor}{RGB}{210,73,42}    % orange

\usepackage{todonotes}

\usepackage{amsmath}
\usepackage{amssymb}
\usepackage{amsthm}
\theoremstyle{definition}

\makeatletter
\let\NAT@parse\undefined
\makeatother
\usepackage[pdfa,colorlinks,bookmarksopen,bookmarksnumbered,allcolors=gg]{hyperref}
\usepackage{cite}
\usepackage[algoruled,vlined,linesnumbered]{algorithm2e}

\SetAlgoSkip{SkipBeforeAndAfter}

\SetCommentSty{mycommfont}
\SetKw{Continue}{continue}
\usepackage{booktabs}
\usepackage{multirow}
\usepackage{makecell}
\usepackage{tabularx}

\usepackage[font=footnotesize]{caption}
\usepackage[font=footnotesize]{subcaption}
\usepackage[export]{adjustbox}

\newcommand{\jeep}{\textsc{Jeep}\xspace}
\newcommand{\van}{\textsc{MercedesVan}\xspace}
\newcommand{\cola}{\textsc{CarlaCola}\xspace}
\newcommand{\vwVan}{\textsc{VolkswagenVan}\xspace}
\newcommand{\method}{\textsc{dgt-map}\xspace}
\newcommand{\baselineSingleTask}{\textsc{dgt-map(Single-Task)}\xspace}
\newcommand{\baselineAverage}{\textsc{dgt-map(Isotropic)}\xspace}
\newcommand{\baselineBinary}{\textsc{Binary}\xspace}
\newcommand{\baselineSlopeBased}{\textsc{Slope-Based}\xspace}
\usepackage[
activate   = {true},
protrusion = true,
expansion  = true,
kerning    = true,
spacing    = true,
tracking   = false,
auto       = true,
selected   = true,
factor     = 1000,
stretch    = 15,
shrink     = 15,
]{microtype}

\title{\LARGE \bf
DGT-Map: Directional Global Traversability Mapping Utilizing Multi-Task Learning for Heterogeneous Vehicles }

\author{Anonymous}

\author{
Jaskrit Singh, Kashif K. Noori, Jing Xiao, and Constantinos Chamzas%
\thanks{All authors are affiliated with the Department of Robotics Engineering , Worcester Polytechnic Institute (WPI), Worcester, MA 01609, USA
{jsingh3, kknoori, jxiao2, cchamzas} @ wpi.edu. We acknowledge the technical and financial support of the Automotive Research Center (ARC) in accordance with Cooperative Agreement W56HZV-24-2-0001 U.S. Army DEVCOM Ground Vehicle Systems Center (GVSC) Warren, MI. DISTRIBUTION STATEMENT A. Approved for public release; distribution is unlimited. OPSEC10544}
}

\begin{document}

\maketitle
\thispagestyle{empty}
\pagestyle{empty}

\begin{abstract}
Off-road traversability is direction-dependent and vehicle specific, yet most global maps assign a single isotropic cost to each location. Existing learned estimators are also commonly trained independently for each vehicle; this preserves vehicle-specific behavior but prevents vehicles from sharing common terrain representations. \method addresses both limitations through a self-supervised framework that learns global, directional, and vehicle-conditioned traversability costmaps from RGB-D observations and locomotion signals. A shared multi-task backbone learns common terrain features across training vehicles while vehicle-specific prediction heads preserve platform-dependent responses. At inference, \method produces a heading-indexed costmap that can be used by a direction-aware planner. We evaluate \method in simulation by integrating it into a Hybrid A* navigation stack and measuring downstream task success on challenging terrains, including slopes that are traversable downhill but not uphill and a ridge obstacle that is traversable by some vehicles, but not by others. Across evaluated tasks, \method achieves the highest or tied-highest navigation success rate when compared against geometric, binary, and learned direction-agnostic baselines.

% Off-road traversability is continuous, direction-dependent, and strongly influenced by vehicle characteristics, yet most global traversability maps assign a single isotropic cost per location, while learned estimators are commonly trained for individual vehicles. \method is a self-supervised framework that learns directional, vehicle-conditioned global traversability costmaps from RGB-D observations by predicting a locomotion-derived signal for terrain patches. A shared multi-task backbone with vehicle-specific prediction heads transfers terrain representations among training vehicles and improves performance for data-scarce vehicles. At inference, \method produces a heading-indexed costmap that can be used by a direction-aware planner. We evaluate \method in simulation by integrating it into a Hybrid A* navigation stack and measuring downstream task success on challenging terrains, including slopes that are traversable downhill but not uphill and a ridge obstacle that is traversable by some vehicles, but not by others. Across tasks and vehicles, \method improves navigation success by 30–40\% over geometric, binary, and learned direction-agnostic baselines.
\end{abstract}

\section{introduction}

Autonomous navigation in unstructured off-road environments remains fundamentally challenging \cite{TraverseSurvey, terra_pn} due to the heterogeneous and geometry-dependent nature of natural terrain \cite{how_does_it_feel,terramechanics}. Unlike structured road networks, where traversable regions are well-defined and largely direction-invariant, off-road terrain exhibits continuous variation in slope, roughness, deformability, and obstacle distribution\mbox{\cite{terramechanics,how_does_it_feel}}
.
%Critically, traversability is not only spatially varying but also direction-dependent: a vehicle that can safely descend a slope may be unable to ascend it due to power limitations.
%As a result, successful navigation requires reasoning not only about \emph{where} traversal is possible, but also about \emph{how the cost and feasibility of traversal vary with direction}.

%A common approach to off-road navigation is to construct a traversability map of the environment and use it to guide a motion planner toward safe and efficient paths~\cite{TraverseSurvey}. However, many existing traversability representations assign a single scalar cost \cite{v_strong, how_does_it_feel} to each spatial location, implicitly assuming that terrain difficulty is isotropic. This simplification discards critical directional effects, such as the difference between ascending and descending a slope as shown in \autoref{fig:intro} top. When used for global planning, such representations limit a planner’s ability to reason about terrain-induced risk and often result in overly conservative or suboptimal paths. \textbf{ thus we want  a map that is directional}

\begin{figure}
    \centering
    \includegraphics[width=1\linewidth]{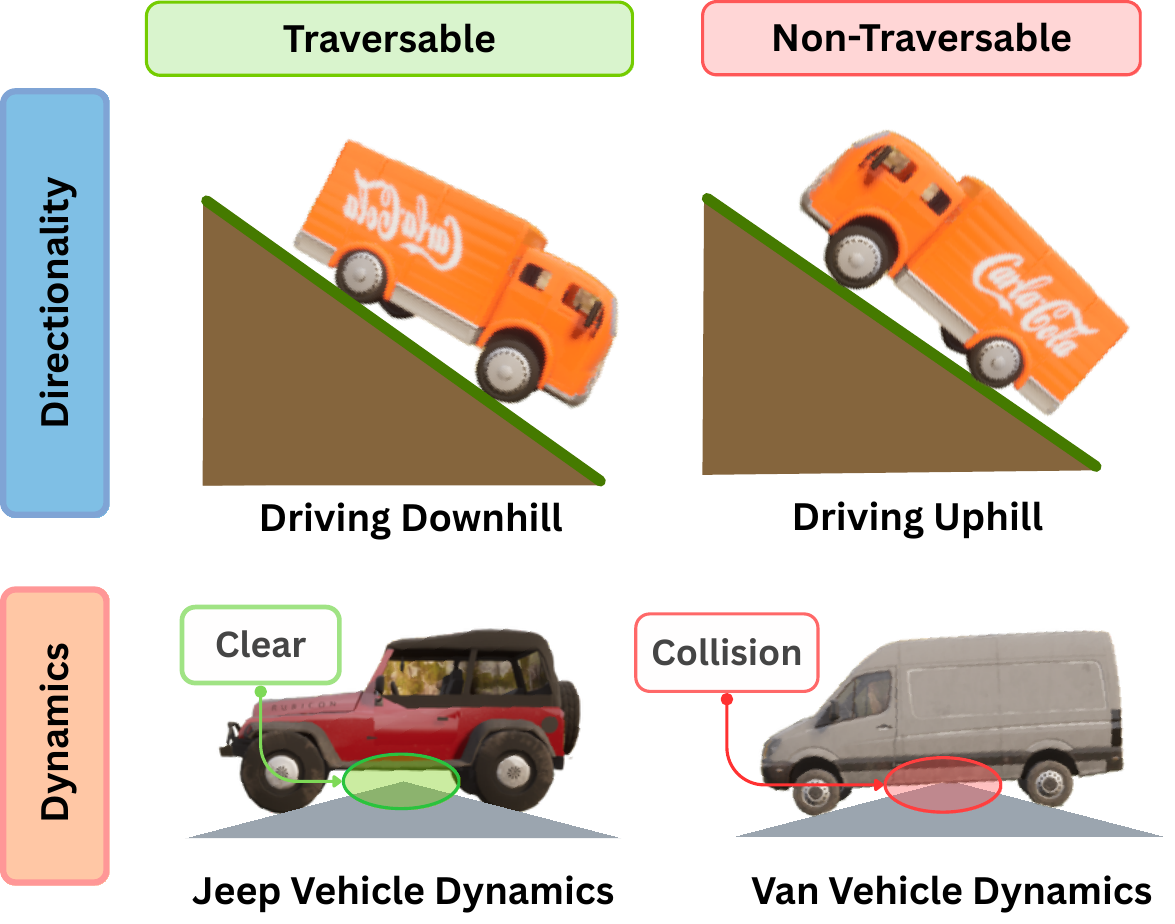}
    \caption{\textbf{Top:} The traversability of terrain depends on the driving direction of the vehicle. For example, a vehicle might be able to go down a hill (left) but not up (right) due to limited horsepower. \textbf{Bottom:} Similarly, the traversability depends on the type of vehicle. For example, vehicles with a long wheel-base, like the van shown on the right, can collide when driving over ridge terrain, while a vehicle with higher ground clearance and a shorter wheel-base can drive over it (left).}  
    \label{fig:intro}
\end{figure}

As a result, a common approach to off-road navigation is to construct a traversability map to guide a navigation planner~\cite{TraverseSurvey}. Most existing methods assign a single scalar cost to each spatial location~\cite{v_strong, how_does_it_feel,terra_pn, unity_sim, how_rough_is_the_path,metaverse, tnt}, effectively modeling traversability as an isotropic property of the terrain. While computationally convenient, this representation collapses directional structure into a single value and prevents the planner from distinguishing between different approach angles at the same location. For example, as shown in \autoref{fig:intro} (top), a vehicle may be able to safely descend a slope but may be unable to ascend it; an isotropic cost map cannot represent this asymmetry. Consequently, planning with such a map often produces overly conservative or suboptimal paths.

%Consequently, the planner is forced to reason over an incomplete model of the environment, which can lead to conservative behavior or suboptimal route selection.

%Paragraph 3 Motivate self-supervision

Beyond directional effects, traversability is inherently vehicle-specific \cite{TraverseSurvey}. The feasibility of crossing a terrain feature depends on vehicle geometry and dynamics, including wheelbase, ground clearance, and power-to-weight ratio. As illustrated in~\autoref{fig:intro} (bottom), a short-wheelbase, high-clearance vehicle (\jeep) can successfully traverse a ridge, whereas a longer vehicle (\van) collides with the same terrain due to geometric constraints. Despite this variability, most existing traversability models are trained for a single vehicle \cite{how_does_it_feel,how_rough_is_the_path,terramechanics, unity_sim}. Applying such models to vehicles with different geometry or dynamics may therefore require retraining or fine-tuning.

% and implicitly assume that the representation transfers across vehicles. This assumption limits generality and necessitates retraining or manual adaptation when deploying on new vehicle types.

Despite substantial progress in traversability estimation \cite{TraverseSurvey}, representations that explicitly model directional or vehicle-dependent traversability remain relatively underexplored. In this work, we present \method, a self-supervised framework that predicts vehicle-conditioned, directional traversability costmaps from RGB-D observations. By training jointly across heterogeneous vehicles, \method allows terrain experience to be shared across vehicles. This is particularly useful when vehicle-specific data is limited, as may occur for large or safety-critical vehicles that are costly to deploy~\cite{caruana1997multitask}.

Concretely, \method constructs a global bird's-eye-view (BEV) map and uses execution feedback to learn locomotion-derived traversability costs for heading-aligned terrain patches \cite{how_does_it_feel,how_rough_is_the_path}. Its multi-task architecture combines a shared terrain encoder with vehicle-specific prediction heads, capturing common terrain structure without treating traversability as identical across vehicles~\cite{caruana1997multitask}.

\method is evaluated by integrating output directional traversability maps into a complete navigation stack (Hybrid A* planning and closed-loop control) and measuring navigation success rates.
Across challenging off-road tasks and multiple vehicle platforms, \method  achieves the highest or tied-highest success-rate when compared against geometric, binary, and direction-agnostic baselines.

The contributions of this work are the following:
\begin{itemize}
    \item A global traversability mapping framework that is \emph{direction-aware} and \emph{vehicle-conditioned}, enabling planners to reason about heading-dependent and vehicle-specific terrain effects.
    \item A self-supervised, multi-task learning formulation that leverages heterogeneous vehicle experience to improve prediction quality for vehicles with limited data.
    \item A closed-loop evaluation in simulation that evaluates the effects of directional representation, learned terrain features, and multi-task learning on navigation success.
\end{itemize}

\section{Related Work}

We organize prior work along two fundamental axes: (i) how traversability is represented, (ii) how it is estimated from perceptual and proprioceptive signals.
%In this section, we review representative approaches along these dimensions and highlight limitations that motivate direction-aware, vehicle-conditioned global traversability mapping.

\subsection{Traversability Representation}
A central question in off-road navigation is how terrain difficulty should be represented for planning. Existing approaches range from discrete class labels to continuous scalar fields and directional cost models.

\textbf{Binary Representations:} Early work formulates traversability as a binary classification problem, labeling terrain as traversable or non-traversable \cite{binary_travers}. Extensions introduce small categorical sets such as safe, risky, and obstacle \cite{geom_seg_travers}. These representations integrate naturally with occupancy-grid planners but collapse terrain variability into coarse categories. In highly heterogeneous off-road environments, such discretization struggles to distinguish between terrain types such as gravel, mud, grass, or deformable soil.

\textbf{Continuous Representations:} To improve expressiveness, many works model traversability as a continuous scalar field over spatial locations \cite{unity_sim,v_strong, how_does_it_feel, how_rough_is_the_path, terra_pn, metaverse, tnt}. Scalar costmaps provide smooth gradients for optimization-based and graph-search planners and are widely adopted in both local and global planning frameworks. However, by assigning a single value to each spatial cell, these methods implicitly assume isotropy. As illustrated in \autoref{fig:intro} (top), terrain such as slopes exhibits strong direction-dependent effects that cannot be captured by an isotropic scalar field.

Some approaches \cite{unity_sim,v_strong, how_does_it_feel, terra_pn, metaverse} partially address this by predicting traversability in the vehicle’s frame of reference. While this allows predicted cost to vary with the vehicle’s heading at inference time, the resulting map still assigns a single cost to each spatial location and therefore cannot fully represent direction-dependent traversability.

% While this allows the predicted cost to vary with the robot’s heading at inference time, the resulting map still assigns a single cost to each spatial location. Consequently, planners cannot assign different costs to the same cell depending on the direction of travel, limiting their ability to explicitly reason about anisotropic terrain structure during search.

\textbf{Directional Representations:} Directional traversability models address this limitation by conditioning cost on heading \cite{terramechanics}. These approaches preserve anisotropic terrain structure and enable planners to reason about ascent versus descent or preferred approach angles. 

While \cite{terramechanics} demonstrates the utility of directional traversability for improving geometric path quality metrics, the evaluation is limited to unconstrained path generation. It does not integrate with a planning and control stack to measure the impact on navigation performance.

\subsection{Traversability Estimation}

Given a traversability representation, the next challenge is estimating traversability from perceptual observations. Existing approaches broadly fall into two categories: geometric feature-based methods and self-supervised methods grounded in collected proprioceptive signals.

\textbf{Geometric feature-based methods:} Classical traversability estimation methods compute hand-engineered terrain descriptors from LiDAR or depth data, such as slope, step height, roughness, or surface curvature \cite{geom, geom_seg_travers, geom_seg_travers_2}. These features are either thresholded directly \cite{geom, geom_seg_travers_2} or fed into a regression model \cite{geom_seg_travers}. Because geometry is inherently directional, these methods can capture effects such as uphill versus downhill traversal. However, purely geometric features struggle to distinguish visually similar geometries with different physical properties (e.g., grass versus mud) and require manually designed mappings from features to cost.

%\textbf{Semantic learning approaches} attempt to address this limitation by incorporating visual information. Some works combine semantic segmentation with geometric reasoning \cite{geom_seg_travers, geom_seg_travers_2}, assigning terrain classes and mapping them to predefined traversability costs. While this improves surface-type discrimination, it introduces reliance on human-annotated labels and hand-designed class-to-cost mappings. Furthermore, semantic regions are typically assigned a single cost, limiting spatial resolution and rarely encoding direction-dependent effects.

%\todo{This and the previous pargraph read well and have sound argument, but we can't off don't show this in our experiment, unluess we claim the different colored terrain means something beyond geometry, which it kind of does not. Hmm lets discuss this a bit }

\textbf{Self-supervised methods:} More recent approaches avoid manually specifying what constitutes good traversability, and instead leverage collected vehicle interaction data and use them as supervisory labels \cite{unity_sim,  how_rough_is_the_path, terra_pn, terramechanics}. In these approaches, perceptual observations of terrain are paired with proprioceptive feedback recorded during traversal, such as vertical acceleration \cite{how_does_it_feel} or discrepancies between commanded and realized motion \cite{unity_sim}. This paradigm grounds traversability in measurable execution outcomes, thereby avoiding manual labeling. Moreover, because the vehicle physically traverses terrain in a specific direction, the resulting supervision naturally captures direction-dependent effects.

%\todo{Discuss with Jing if we should have a pargraph like this, instead of the motivation}
Despite their advantages, most self-supervised approaches are trained independently for a single vehicle. Since traversability depends strongly on vehicle geometry and dynamic capabilities, models trained on one vehicle often require retraining or fine-tuning when deployed on another. To the best of our knowledge, multi-vehicle self-supervised training for traversability estimation remains largely unexplored.

\section{Problem Statement}
\label{sec:problem_statement}
We consider the problem of learning a \emph{global, directional, vehicle-conditioned traversability map} for off-road navigation using self-supervised data. 
Given $N$ vehicles equipped with RGB-D cameras and IMUs operating in an unknown off-road environment, our goal is to estimate terrain-induced traversal cost as a function of spatial location, direction of travel, and vehicle platform.

Formally, we seek to learn a traversability function
\begin{equation}
    c : \mathbb{R}^2 \times \Theta \times \mathcal{V} \rightarrow [0,1],
    \label{eq:travesability_formulation}
\end{equation}
where $c(x, y, \theta, v)$ denotes the expected cost of traversing location $(x, y)$ with heading $\theta \in \Theta = [0, 2\pi]$ for vehicle $v \in \mathcal{V}$. Here $\mathcal{V}$ denotes the set of $N$ vehicles represented during training. The cost reflects terrain-induced vehicle disturbance and is defined such that higher values correspond to more difficult or undesirable traversal.

We evaluate the learned traversability function by integrating it with a navigation system and measuring the task success rate in navigating to a given goal.

\section{Methodology}

Our approach learns the traversability function defined in \autoref{sec:problem_statement} through self-supervision: rather than manually labeling terrain, we use the discrepancy between commanded and actual vehicle motion as a natural measure of traversal difficulty (\autoref{sec:data-collection}). We then use a multi-task CNN with a shared terrain backbone and vehicle-specific prediction heads, allowing terrain features to be shared across vehicles while preserving vehicle-dependent responses (\autoref{sec:training}). After training our traversability model, we can use it to predict a traversability map for a new environment (\autoref{sec:costmap-inference}). We can then run a traversability-aware planner (\autoref{sec:planning}) on the map to generate safe paths.

% . We also found that this shared training formulation stabilizes learning from the noisy self-supervised signal
% Since this produces a noisy training signal, we found it beneficial to utilize a multi-task Convolutional Neural Network (CNN) trained on multiple vehicles simultaneously to stabilize the training process (\autoref{sec:training}).

% We train a multi-task CNN to predict traversability across four different vehicles, allowing 

% To capture these dependencies without manual annotation, we adopt a self-supervised pipeline with three stages. First we collect data by driving vehicles through diverse terrain and measuring measuring the discrepancy between the command and actual motion. Next
% \begin{itemize}
%     \item \textbf{A) Data Collection} where vehicles are driven through diverse terrain and the discrepancy between commanded and achieved motion provides a traversability signal requiring no human labeling
% \end{itemize}
% data collection, ; (2) model training, where a multi-task CNN learns to predict this signal from terrain appearance across multiple vehicle platforms; and (3) planning, where the trained model generates a directional costmap that a Hybrid A* planner queries to produce heading-aware, vehicle-specific paths.

\subsection{Data Collection}
\label{sec:data-collection}

\begin{figure*}[t]
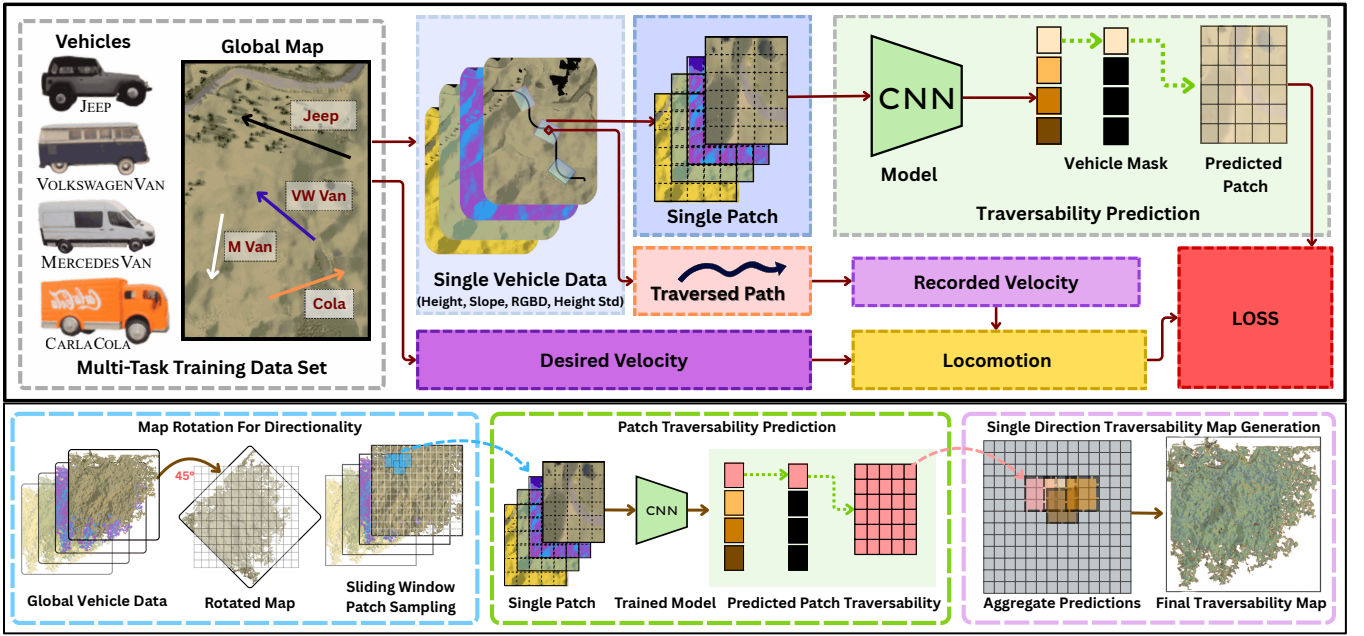

    \centering
    \includegraphics[width=\linewidth]{figures/Methodology/Training-1.pdf}
    \vspace{0.75em}
    \includegraphics[width=\linewidth]{figures/Methodology/Testing_small.pdf}
    \caption{
    \textbf{DGT-Map training and inference pipeline.}
    \textbf{Top:} Multi-vehicle self-supervised training. RGB-D observations from multiple vehicles are fused into a global BEV map, from which heading-aligned terrain patches are extracted along the traversed path. Locomotion labels are computed from the discrepancy between desired and recorded motion. A shared CNN is trained to predict these labels from terrain patches, with vehicle-specific heads preserving platform-dependent traversability behavior. Joint training across vehicles enables shared terrain representation learning.
    \textbf{Bottom:} Directional global traversability map generation at inference. The global BEV map is rotated to align a queried heading with the network’s forward axis. A sliding window extracts patches, which are processed by the trained CNN, and overlapping predictions are aggregated to form a costmap for a single-direction. Repeating this procedure across headings yields the full directional traversability map used for planning.
    }
    \label{fig:full_pipeline}
\end{figure*}

Self-supervised traversability learning requires associating terrain appearance with vehicle-terrain interaction outcomes. Although failed traversal attempts provide useful examples of difficult terrain, less capable vehicles may become immobilized before exploring large portions of challenging environments, limiting the diversity of terrain they observe. To broaden coverage, we use four geometrically and dynamically distinct vehicles, shown in \autoref{fig:full_pipeline} (top-left): \jeep, \vwVan, \van, and \cola. Of these, the \jeep can traverse the widest range of terrain and therefore contributes experiences from regions that may be inaccessible to the less capable vehicles. 

% Self-supervised traversability learning requires associating terrain appearance with vehicle-terrain interaction outcomes. However, since it can be difficult for less capable vehicles to collect diverse data in challenging terrain, we drive 4 geometrically different vehicles to cover a larger range of different types of terrains. These vehicles 
% % Since a vehicle must traverse a piece of terrain to collect training data, the untraversable part of the terrain we are interested in modeling will not be part of the collected data. This issue is addressed by driving four geometrically and dynamically distinct vehicles 
%  are shown in \autoref{fig:full_pipeline}, top-left; from top to bottom these are \jeep, \vwVan, \van, \cola. Of these, the \jeep can traverse the most terrain, providing an expanded set of experiences that other vehicles can utilize.
Data collection begins by constructing a global map of the training environment. Terrain patches are then extracted along each run trajectory and paired with a locomotion signal that reflects the terrain-induced difficulty experienced during traversal. The resulting patch--locomotion pairs form the self-supervised training dataset.

% Data collection begins by constructing a global map of the training environment. Terrain patches are then extracted along each run trajectory and paired with their, to create the training dataset. 
% First we aggregate the RGB-D data from all the runs to create a global Birds Eye View (BEV) map of the training environment \autoref{sec:global_map}. Next we create training samples for each run. We first take BEV patches from the global map along the path of the run \autoref{sec:patches}. Then we calculate locomotion along the path as our traversability signal \autoref{sec:locomotion}. Finally we associate patches to their locomotion values to create training sample we can feed into our model \autoref{sec:training-samples}.

% Each vehicle is spawned at random positions on the map and driven forward with speed regulated by a PID controller. Multiple runs are accumulated to build a dense global Birds Eye View (BEV) map. We respawn in a random location after 100 seconds or when the vehicle gets stuck, whichever happens first. 

% \subsection{Pre-filtering:}
% \label{sec:pre-filtering}
% Frames with extremely high angular momentum (greater than $100^o /sec$) are removed. These events correspond to simulator glitches where the vehicle is launched into the air after a collision.

% We also correct for the camera’s vignette effect, which darkens the corners of the image and introduces artifacts in the BEV map. To remove this effect, we calibrate the camera using an image of a uniformly white floor.

\subsubsection{Global Map}
\label{sec:global_map}
Individual RGB-D frames provide only a partial, viewpoint-dependent snapshot of the environment. By aggregating depth and color observations across all runs into a single spatial grid at 0.4 m resolution, we obtain a terrain representation that is complete and independent of any single vehicle trajectory.

The aggregate RGB-D data forms a bird's-eye-view (BEV) map with six channels:
\[
R,\; G,\; B,\; \bar{H},\; \sigma_H,\; \|\nabla \bar{H}\|.
\]

Here, $R, G, B$ are color channels, $\bar{H}$ is the average height per cell, $\sigma_H$ is the standard deviation of the height values within each cell, and $\nabla \bar{H}$ is the gradient of $\bar{H}$ computed using central differences over the four neighboring cells.

\subsubsection{Patches}
\label{sec:patches}

Because the CNN must learn direction-dependent traversability, it must always see terrain from a consistent viewpoint relative to the vehicle's heading. We achieve this by extracting patches in the vehicle's local frame, with the forward direction aligned with the top of the patch (\autoref{fig:full_pipeline}, top-center). This heading-aligned extraction later enables us to create directional traversability maps.

Each patch measures 8.8 m × 2.8 m, sized to fully cover all vehicle footprints while including surrounding terrain context.

\subsubsection{Locomotion}
\label{sec:locomotion}
With terrain patches in hand, we need a supervision signal that reflects the difficulty of traversing the terrain. Locomotion provides a continuous, scalar measure of how closely the vehicle's realized motion matches its commanded motion over a terrain segment. Similar to \cite{terramechanics}, we define locomotion as:
\begin{equation}
\label{eq:locomotion}
    L = \exp\left[
-\frac{1}{2\sigma^2}
\left(\frac{u\Delta t - \|\Delta \mathbf{x}\|}{u\Delta t}\right)^2
\right],
\end{equation}
where $L \in [0,1]$, $\Delta \mathbf{x}$ is the vehicle displacement over interval~$\Delta t$, and $u = 5\,\text{m/s}$ is the target velocity maintained by a PID controller during data collection. We set $\sigma = \tfrac{1}{3}$ following~\cite{terramechanics}. 

Higher values of $L$ indicate that the vehicle more closely maintained its commanded motion and therefore correspond to easier traversal, while lower values indicate greater terrain-induced difficulty. Thus, locomotion has the opposite convention from the traversability cost used later in Sec.~\ref{sec:costmap-inference}. 

\subsubsection{Self-Supervised Label Generation}
\label{sec:training-samples}
The locomotion equation is applied in a sliding window over each run for vehicle $v$, producing a time-indexed signal $L_v(t)$. Each patch $P$ extracted at time $t_P$ is paired with $L_v(t_P)$, yielding a dataset of patch--locomotion pairs across all vehicles. Because different vehicles may traverse the same terrain and yield different outcomes, this dataset naturally encodes both terrain-dependent and vehicle-dependent variation.

\subsection{Training}
\label{sec:training}

The dataset produced above pairs terrain appearance with a vehicle-specific traversability signal, but contains an inherent imbalance: some vehicles (e.g., \jeep) contribute far more training patches than others (e.g., \cola) due to differences in traversal range. Rather than training separate models for each vehicle, we adopt a multi-task formulation~\cite{caruana1997multitask} that jointly trains on data from all vehicles while retaining vehicle-specific outputs.

\subsubsection{Multi-Task Learning}

The model consists of a shared feature backbone followed by $N$ vehicle-specific regression heads, one per vehicle. For a patch collected by vehicle $v$, the network produces $N$ locomotion predictions $\{\hat{L}_1(P), \dots, \hat{L}_N(P)\}$, but the loss is computed only for the head corresponding to vehicle $v$. The remaining outputs are masked and do not contribute to the gradient.

\begin{equation}
    \mathcal{L}(P, v) = 
    \left\| \hat{L}_v(P) - L_v(t_P) \right\|^2.
\end{equation}

This design allows the backbone to learn a shared terrain representation while the individual heads capture vehicle-specific responses. Sharing the backbone also regularizes training and allows lower-data vehicles to benefit from representation learning over the broader multi-vehicle dataset.

% \label{sec:training}
% The dataset produced above pairs terrain appearance with a vehicle-specific traversability signal, but contains an inherent imbalance: some vehicles (e.g., \jeep) contribute far more training patches than others (e.g., \cola) due to differences in traversal range. Rather than training separate models for each vehicle, we adopt a multi-task formulation~\cite{caruana1997multitask} that pools data across vehicles to learn a shared terrain representation while retaining vehicle-specific outputs. This allows vehicles with less training data to benefit from terrain experience collected by other platforms without assuming that traversability is identical across vehicles.

% \subsubsection{Multi-Task Learning}

% The model consists of a shared feature backbone followed by $N$ vehicle-specific regression heads, one per vehicle. For a patch collected by vehicle $v$, the network produces $N$ locomotion predictions $\{\hat{L}_1(P), \dots, \hat{L}_N(P)\}$, but the loss is computed only for the head corresponding to vehicle $v$. The remaining outputs are masked and do not contribute to the gradient.

% \begin{equation}
%     \mathcal{L}(P, v) = 
%     \left\| \hat{L}_v(P) - L_v(t_P) \right\|^2.
% \end{equation}

% This design enables the backbone to learn terrain features that are shared across vehicles, such as roughness patterns or slope profiles, while the heads learn how each vehicle responds to these features. The shared backbone regularizes training and allows lower-data vehicles to benefit from terrain representations learned from the broader multi-vehicle dataset.

\subsubsection{Network Architecture}
The six BEV channels naturally divide into two groups with distinct statistical properties: RGB color (which benefits from pretrained visual features) and height statistics ($\bar{H},\; \sigma_H,\; \|\nabla \bar{H}\|$), which have no natural analog in standard image datasets. The network reflects this split with a dual-branch architecture that uses two ResNet-18 backbones: the RGB branch is initialized with ImageNet weights, and the geometric branch is trained from scratch. Their feature embeddings are concatenated and then passed through a small fully connected network before being fed into the vehicle-specific regression heads.

\subsection{Costmap Inference}
\label{sec:costmap-inference}

The trained model predicts locomotion for a single forward-facing patch, but planning requires a dense cost estimate at every spatial location and for every travel direction. We bridge this gap with the costmap inference pipeline shown in \autoref{fig:full_pipeline} (bottom).

% At inference time, the learned model is used to construct a global directional traversability map approximating the continuous function 
% $c(x,y,\theta,v)$ from \autoref{sec:problem_statement}. 
% The overall procedure is illustrated in \autoref{fig:full_pipeline} (bottom).

\subsubsection{Spatial and Angular Discretization}

The global BEV map is represented as a spatial grid indexed by $(i,j)$. 
To capture direction-dependent effects, orientation is discretized into K bins:
\begin{equation}
\theta_k = k \cdot \frac{2\pi}{K}, \quad k \in ~\{0, \dots, K-1\}.
\end{equation}

This yields a discrete representation over $(i,j,\theta_k)$ for each vehicle $v$. For experiments, we used $K=8$.

\subsubsection{Patch Evaluation and Costmap Construction}

A sliding window approach (shown in \autoref{fig:full_pipeline}
 (bottom)) is used to convert patch predictions into a costmap. When patches overlap, predicted costs are averaged over all patches covering a cell to produce a dense estimate. Since the model predicts locomotion exclusively in the forward direction, the costmap specifically represents traversability with heading $\theta = \theta_0 =  0$. 
 
 Costmaps for other heading $k \in ~\{1, \dots, K-1\}$ can be generated by rotating the map by $-\theta_k$ before feeding it to the CNN with the sliding window approach.

% The key insight is that the network was trained exclusively on forward-facing patches, so it only knows how to predict traversability for a single heading. We can use this to predict a traversability map for all headings, $\theta_k$, by rotating the gl

% Rather than retraining for every direction, we rotate the entire BEV map so that the desired heading $theta_k$ aligns with the network's forward axis (\autoref{fig:full_pipeline}, bottom-left).

% For each heading $\theta_k$, the global BEV map is rotated so that $\theta_k$ aligns with the network’s forward axis (see \autoref{fig:full_pipeline}, bottom-left). This allows the network to make predictions for a vehicle driving at heading $\theta_k$.

% A sliding-window procedure then extracts patches of the same size used during training. Each patch centered at grid cell $(i,j)$ is processed independently by the CNN to predict locomotion values $L(i,j,\theta_k,v)$ for each vehicle.

Finally, the output of the network is converted from locomotion to cost via
\begin{equation}
C(i,j,\theta_k,v) = 1 - L(i,j,\theta_k,v).
\end{equation}

%  Repeating this procedure for all $K$ orientations yields a multi-layer directional costmap,
% \begin{equation}
% C(i,j,\theta_k,v) \approx c(x_i,y_j,\theta_k,v).
% \end{equation}

Examples of directional layers are shown in \autoref{fig:planning}(c).

% \subsubsection{Cost Inflation}

% In practice the controller does not track the planned path perfectly. To maintain a safety margin, high-cost regions are inflated by assigning each cell the maximum cost within a spatial neighborhood of radius~$R$:
% \todo{what is the value of R used?}
% %
% \begin{equation}
%     C_{\text{inflated}}(i,j,\theta_k,v)
%     =
%     \max_{(u,w)\in \mathcal{N}_R(i,j)}
%     C(u,w,\theta_k,v).
% \end{equation}

%%[Kashif]
\subsection{Traversability-Aware Path Planning}
\label{sec:planning}
Now that we have a directional costmap, we can integrate it with a traversability-aware planner.

The directional costmap encodes \emph{where} terrain is difficult and \emph{from which direction}, but it does not distinguish between terrain that is merely costly and terrain that is physically impassable. A steep slope may have a high traversability cost in one direction while remaining feasible, whereas a cliff is impassable regardless. The planner therefore operates on two decoupled map layers: a binary occupancy map for hard-collision checking and a directional costmap for soft-traversability preferences.
% After generating our directional costmap we can use it to plan low-cost paths. While planning, we select the appropriate costmap layer based on the direction of travel at each expansion as shown in \autoref{fig:planning}. We integrate this lookup into a Hybrid~A* planner~\cite{dolgov2010path, bjelland2021path}.

%, but our costmap could also be used with other planners such as ~\cite{koenig2002dstarlite, ferguson2006field, daniel2010theta}. 

% \subsubsection{Traversability-Aware Path Planning}
% \label{sec:planning}

% The directional costmap constructed in \autoref{sec:costmap-inference} provides a heading-dependent cost function $c(x,y,\theta_k)$ over the workspace. 
% Unlike standard grid-based planners, which assign a single scalar cost to each cell and therefore discard directional information, our representation maintains multiple cost layers corresponding to discretized headings.

% ---------- A. Occupancy Map ----------
\subsubsection{Occupancy Map}
\label{sec:occupancy_map}
%  We construct the occupancy map using the height gradient, $\|\nabla \bar{H}\|$, from the BEV map (\autoref{sec:global_map}).  If the slope exceeds a threshold~$\tau_s$, the cell is marked as occupied:
% %
% \begin{equation}
%     \mathrm{occ}(i) =
%     \begin{cases}
%         1 & \text{if } \|\nabla \bar{H}(i)\| > \tau_s, \\
%         2 & \text{if cell } i \text{ is unobserved,} \\
%         0 & \text{otherwise.}
%     \end{cases}
%     \label{eq:occupancy}
% \end{equation}
% %
% This geometric rule captures hard obstacles such as cliffs, trees, and steep embankments. We select $\tau_s$ such that it is the smallest value that does not eliminate terrain that is traversable in some direction (such as the slope in \autoref{fig:intro}(top)). \autoref{fig:planning}(b) shows an example of a generated occupancy map.

 The occupancy map is constructed from the height gradient of the BEV map as follows:
\begin{equation}
    \mathrm{occ}(i) =
    \begin{cases}
        1 & \text{if } \|\nabla \bar{H}(i)\| > \tau_s, \\
        2 & \text{if cell } i \text{ is unobserved}, \\
        0 & \text{otherwise}.
    \end{cases}
    \label{eq:occupancy}
\end{equation}
 where $\tau_s$ is selected as the smallest value that does not eliminate terrain traversable in at least one direction, such as the slope in \autoref{fig:intro}(top), which is steep but traversable downhill. \autoref{fig:planning}(b) shows an example of an occupancy map. Unobserved cells are treated as occupied during collision checking.

\subsubsection{Directional Cost Lookup}
\label{sec:cost_lookup}
When expanding from grid cell $(i,j)$ to a successor cell $(i',j')$, 
the planner computes the travel direction
\begin{equation}
    \theta' = \mathrm{atan2}(j' - j,\; i' - i),
    \label{eq:travel_dir}
\end{equation}
where $(i,j)$ and $(i',j')$ denote grid indices.

The nearest discretized heading bin is then selected
\begin{equation}
    k^* = \arg\min_k \left| \theta_k - \theta' \right|.
\end{equation}

\begin{figure}
    \centering
    \includegraphics[width=1\linewidth]{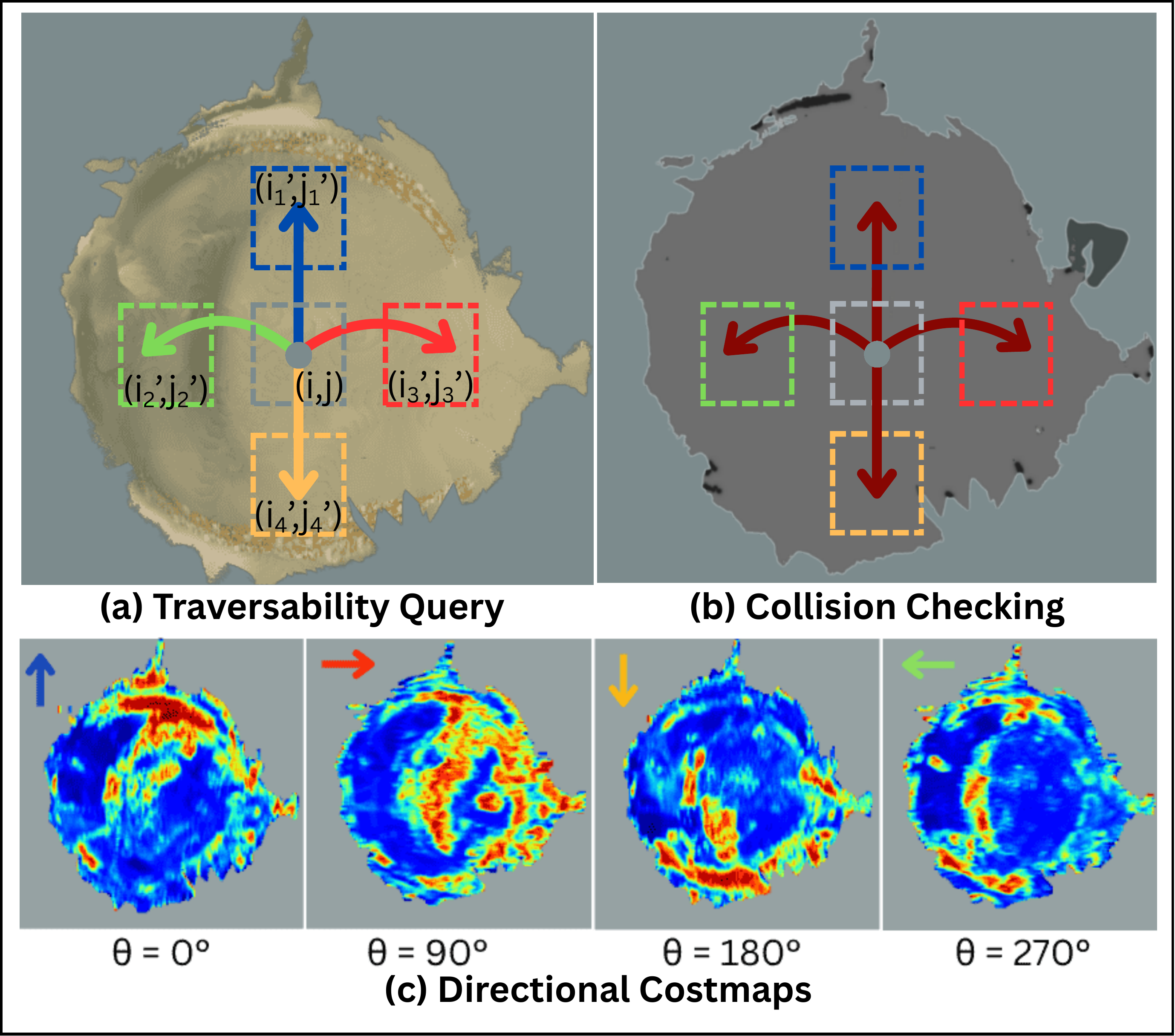}
    \caption{\textbf{Traversability-aware planning.}
    \textbf{(a)} When expanding from grid cell $(i,j)$, the planner computes the travel direction to neighboring cells $(i'_{n}, j'_{n})$ for $n={1,2,3,4}$ and queries the corresponding directional cost layer.
    \textbf{(b)} Each candidate motion primitive is discretized and checked against the occupancy map to ensure collision-free feasibility.
    \textbf{(c)} Four of the $K=8$ directional costmap layers ($\theta = 0^\circ, 90^\circ, 180^\circ, 270^\circ$) illustrating how traversability varies with heading; differences across layers capture anisotropic terrain effects such as ascending versus descending slopes.}
    \label{fig:planning}
\end{figure}

This process is illustrated in \autoref{fig:planning}(a), where different expansion directions query different directional cost layers.

% To reduce discretization artifacts and account for controller tracking uncertainty, 
% we interpolate between the nearest orientation bin and its two angular neighbors:
% \begin{equation}
% \label{eq:dir_blend}
% \begin{aligned}
% \bar{C}(i',j',\theta',v)
% &=
% w_0\, C(i',j',\theta_{k^*},v)
% +
% w_1\, C(i',j',\theta_{k^*-1},v) \\
% &\quad +
% w_1\, C(i',j',\theta_{k^*+1},v).
% \end{aligned}
% \end{equation}
% We use $w_0 = 0.7$ and $w_1 = 0.15$, which provide smooth angular transitions while preserving directional contrast.

Finally, a nonlinear amplification is applied to penalize high-cost terrain more aggressively than low-cost terrain:
\begin{equation}
C'(i',j',\theta',v)
=
C(i',j',\theta_{k^*},v)
\left(1 + \beta\, C(i',j',\theta_{k^*},v)\right),
\label{eq:stretch}
\end{equation}
where $\beta \ge 0$ controls the strength of the super-linear amplification. The value of $\beta$ determines the tradeoff between path length and terrain traversability. Experimentally, $\beta=150$ provided sufficient penalty to avoid high-cost areas without causing excessive detours around minor terrain irregularities.

% Then a nonlinear amplification is applied to penalize high-cost terrain more aggressively than low-cost terrain:

% \begin{equation}
% C'(i',j',\theta',v)
% =
% C(i',j',\theta_{k^*},v)
% \left(1 + \beta\, C(i',j',\theta_{k^*},v)\right),
% \label{eq:stretch}
% \end{equation}
% where $\beta \ge 0$ controls the degree of super-linear amplification. $\beta$ is a parameter that must be tuned based on the desired tradeoff between path length and path traversability. Experimental validation showed that a value of $\beta=150$ resulted in paths that avoided obvious high risk areas while not creating paths that were overly convoluted to avoid minor bumps in the terrain.

\subsubsection{Hybrid A*}
\label{sec:hybrid_astar}
With both map layers defined, we plan using Hybrid~A*~\cite{dolgov2010path, bjelland2021path}, 
which searches over continuous states $(x,y,\theta)$ using arc-based motion primitives 
that respect the vehicle’s minimum turning radius $r_{\min}$. At each node, three motion primitives (straight, left, right) are evaluated, 
each advancing the state by a fixed step size $s$ along an arc of curvature 
$\kappa \in \{0, +1/r_{\min}, -1/r_{\min}\}$.

% For each candidate arc, the path segment is discretized into fine sub-steps 
% and all intermediate cells are checked against the occupancy map 
% (\autoref{fig:planning}(b)). 
% If any cell is occupied, the primitive is rejected.

% The terrain cost is evaluated only at the successor cell (as shown in \autoref{fig:planning}(a)) using 
% the directional lookup. 
% The edge cost is defined as
% \begin{equation}
% g_{\mathrm{edge}}((i,j),(i',j'))
% =
% s \left( 1 + C'(i',j',\theta_{ij},v) \right).
% \label{eq:edge_cost}
% \end{equation}

Collision checking and cost evaluation serve complementary roles during expansion. For each candidate arc, the path segment is discretized into fine sub-steps, and all intermediate cells are tested against the occupancy map (\autoref{fig:planning}(b)). If any cell is occupied, the primitive is rejected. 

The traversability cost, by contrast, is evaluated only at the successor cell using the directional lookup, yielding an edge cost:

\begin{equation}
g_{\mathrm{edge}}((i,j),(i',j'))
=
s \left( 1 + C'(i',j',\theta',v) \right).
\label{eq:edge_cost}
\end{equation}

Every expansion thus incurs at least a distance cost of our step size $s$, with terrain difficulty adding proportionally. This formulation naturally trades off path length against traversability: the planner will take a longer route only if the terrain cost savings justify the additional distance.

We use $s=1.6 m$, which provided a practical tradeoff between path resolution and planning time in our simulation experiments.

% % ---------- C. Hybrid A* ----------
% \subsubsection{Hybrid A*}
% \label{sec:hybrid_astar}
% We plan with Hybrid~A*~\cite{dolgov2010path, bjelland2021path}, which searches over continuous states $(x, y, \theta)$ using arc-based motion primitives that respect the vehicle's minimum turning radius~$r_{\min}$.  At each node we evaluate three primitives (straight, turn left, turn right), each advancing the state by a fixed step size~$s$ along an arc of curvature $\kappa \in \{0,\; +1/r_{\min},\; -1/r_{\min}\}$.

% For each candidate arc, we discretize the segment into fine sub-steps and test every intermediate cell against the occupancy map, as shown in ~\autoref{fig:planning}(b). If any cell is occupied, the arc is rejected. The traversability cost is evaluated only at the successor cell using the directional lookup from Section~\ref{sec:cost_lookup}. The per-edge cost is
% %
% \begin{equation}
%     g_{\mathrm{edge}}(i, j)
%         = s \cdot \bigl(1 + c'(j,\; \theta_{i \to j})\bigr).
%     \label{eq:edge_cost}
% \end{equation}
% %
% So every expansion incurs at least a distance cost of~$s$, with terrain difficulty adding proportionally. The total cost of a path~$\pi$ is
% %
% \begin{equation}
%     J(\pi) = \sum_{(i,j) \in \pi} g_{\mathrm{edge}}(i, j).
%     \label{eq:total_cost}
% \end{equation}
% For the search heuristic, we use Euclidean distance, which gives an admissible lower bound on the direction-dependent cost at each cell.
% % 

%%[Kashif]
\subsubsection{Controller Integration}
\label{sec:exp_setup}
Given a learned directional costmap for a target vehicle, the Hybrid A* planner (Section~\ref{sec:hybrid_astar}) produces a global reference path. 
This path is tracked using the Nav2~\cite{macenski2020marathon2} MPPI controller~\cite{mppi}.

Nav2 MPPI outputs commanded linear and angular velocities \((v_x,\omega_z)\). To interface these commands with CARLA's Ackermann steering controls, we convert them to a steering angle using the kinematic bicycle relation
\begin{equation}
    \delta = \arctan\!\left(\frac{\ell \, \omega_z}{v_x}\right),
    \label{eq:steering}
\end{equation}
where $\ell$ denotes the vehicle wheelbase. The steering command is normalized by the vehicle’s maximum steering angle before being applied. Longitudinal motion is regulated by a PID controller with a feedforward term that compensates for rolling resistance and aerodynamic drag.

\section{Experiments}

% In this section, we empirically evaluate four central aspects of \method:
% \begin{enumerate}
%     \item the effect of retaining directional traversability information;
%     \item the ability to represent vehicle-specific traversability;
%     \item the ability of learned features to capture terrain effects beyond a slope-based model;
%     \item and the benefit of multi-task training relative to single-task training.
% \end{enumerate}

In this section, we empirically evaluate four central claims of our method:
\begin{enumerate}
    \item \method captures global, directional traversability information not captured by direction-agnostic methods.
    \item \method captures vehicle-specific nuances in traversability.
    \item \method captures features beyond basic geometric slope cues.
    \item Multi-task training improves performance compared to single-task training with the same amount of vehicle-specific data, especially in cases where vehicle-specific data is limited.
\end{enumerate}

To evaluate these claims, we design three controlled navigation experiments.

\textbf{Dataset:} 
The collected dataset consists of 30,000 patches from 400 \jeep trajectories, 10,000 patches from 400 \vwVan trajectories, 
10,000 patches from the 300 \van trajectories, and 2,000 patches from the 200 \cola trajectories. Note that some vehicles such as \cola got stuck more frequently, so they got fewer patches per trajectory due to early termination of runs. This data was collected on a training map with hilly terrain and occasional trees and paths similar to the terrain shown in \autoref{fig:full_pipeline}.

\textbf{Experimental Protocol:} 
The data for each vehicle was partitioned into 70\% train, 15\% validation, 15\% test, with all patches from a given trajectory in the same partition. This prevents temporally and spatially correlated patches from the same trajectory from appearing across splits. Additionally, navigation evaluation maps were spatially distinct from the training map. Lastly, all compared methods used identical start–goal queries, occupancy maps, Hybrid A* settings, and
controller settings.

\begin{figure}
    \centering
    \includegraphics[width=1.0\linewidth]{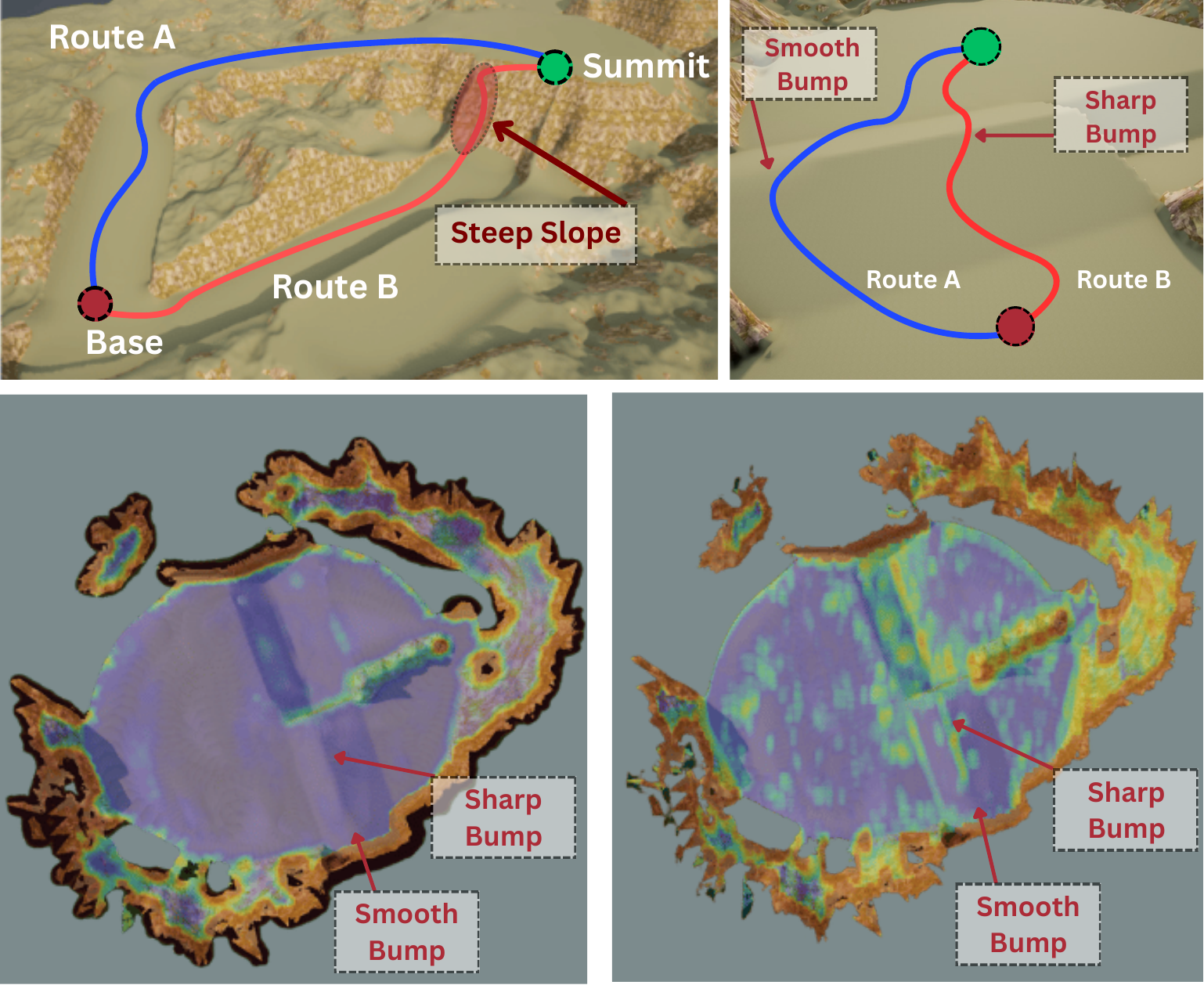}
    \caption{\textbf{Top Left:} \emph{Hill Map} showing Route A, a roundabout route with a gentle slope, and Route B, a direct route containing a steep section. \textbf{Top Right:} \emph{Ridge Map} showing Route A, a path that avoids the sharp ridge, and Route B, a path that crosses the ridge. Route B is not traversable by some vehicles. \textbf{Bottom Left:} Traversability map of the \jeep, note how the ridge is shown as traversable. \textbf{Bottom Right:} Traversability map for the \van, note how the ridge has a low traversability.} 
    \vspace{-1em}
    \label{fig:sharp_hill_combined}
\end{figure}

\textbf{Experiment 1 (Directional Global Traversability):}
To evaluate Claim 1, we use the hill environment shown in \autoref{fig:sharp_hill_combined} (top left). The map contains two feasible routes between the base and summit: 
Route A (long and gentle) and Route B (short with a steep section). The steep segment is designed such that the \cola can safely descend but cannot ascend it.

For each trial, the planner is given only a start and goal location and must compute a complete path over the entire map. No constraints are imposed on route selection. 

A direction-aware global costmap should cause the planner to select Route A during ascent and Route B during descent. Direction-agnostic methods are expected to select the same route in both directions, leading to failures during uphill traversal or overly conservative descents.

\textbf{Experiment 2 (Vehicle-Specific Traversability and Traversability Beyond Slope):}
To evaluate Claims 2 and 3, we use the ridge environment shown in \autoref{fig:sharp_hill_combined} (top right). The ridge can be safely traversed by the \jeep but immobilizes the \van. A smoothed section provides a feasible crossing point for the \van.

As in Experiment 1, the planner must compute a full path from start to goal without prior knowledge of the safe crossing location. This setup tests whether the learned model captures both vehicle-dependent differences and terrain features beyond simple geometric slope cues.

\textbf{Experiment 3 (Multi-Task Training):}
To evaluate Claim 4, we compare \method to its single-task variant, \baselineSingleTask (see \autoref{sec:baselines}). We evaluate both models on random start-and-goal navigation tasks for the \jeep and \vwVan, with the \vwVan providing a lower-data case.

% To evaluate Claim 4, we compare \method with its single-task variant on the \jeep and \vwVan. \vwVan provides a lower-data case, with 10,000 training patches compared with 30,000 for the \jeep.

%we test whether multi-task learning improves performance 
%for vehicles with limited training data. In particular, we focus on the 
%data-scarce \vwVan, which has significantly fewer training patches than the \jeep.

For this experiment, evaluation is performed on a map with terrain similar to the training environment (hilly with occasional trees and paths). 
For each trial, a vehicle is assigned random start and goal locations, and performance 
is measured using the task success rate over multiple runs.

% ============================================================================
% BASELINES
% ============================================================================

\subsection{Baselines}
\label{sec:baselines}
To evaluate these claims, we compare \method against 4 baselines:
\begin{itemize}
    \item \baselineAverage: This method predicts all 8 directional maps using \method, then averages the costmaps, producing a single isotropic costmap.
    \item \baselineSingleTask: This method uses the same architecture as \method, but is only trained on the training data from a single vehicle.
    \item \baselineBinary: This method is not given a costmap, it is only given an occupancy map.
    \item \baselineSlopeBased: This method is engineered based on the assumption that locomotion is correlated with the slope of the ground in the forward direction, i.e., driving uphill is harder than driving on a level surface, (see \autoref{eq:slope-based-predictor}). First, a plane is fitted to the $\Bar{H}$ channel of a terrain patch and the slope is measured in the forward direction, $\theta$. The method assumes there is an angle, $\theta_0$ at which the terrain is fully traversable, which is assigned $L_{sb}(\theta_0)=1$. The method further assumes that increasing or decreasing the slope will cause the locomotion to decrease until it reaches 0, yielding $L_{sb}(\theta_{min})=L_{sb}(\theta_{max})=0$ with $\theta_{min} \leq \theta_0 \leq \theta_{max}$. Furthermore, decay rates are allowed to differ between uphill, $p_u$, and downhill, $p_d$. Finally, the parameters of the model $\{\theta_0, \theta_{min}, \theta_{max}, p_u, p_d\}$ are fit to a given vehicle's training data using Optuna \cite{optuna_2019}.
    \begin{equation}
        L_{sb}(\theta) = \begin{cases}
  1 - \left( \frac{\theta-\theta_0}{\theta_{max}-\theta_0}\right)^{p_u} & \theta \in [\theta_0,\theta_{max}] \\
  1 - \left( \frac{\theta_0-\theta}{\theta_{0}-\theta_{min}}\right)^{p_d} & \theta \in [\theta_{min},\theta_0] \\
  0 & \text{otherwise}
\end{cases}
\label{eq:slope-based-predictor}
    \end{equation}
    As in \autoref{sec:costmap-inference}, locomotion is converted to terrain cost via the formula $c_{sb} = 1 - L_{sb}$.
    
\end{itemize}

% ============================================================================
% RESULTS
% ============================================================================

\subsection{Results}

\begin{table}[t]
  \renewcommand{\arraystretch}{1.35}
  \setlength{\tabcolsep}{3.0pt}
  \centering
  \caption{%
    \textbf{Navigation Success Rate} on two test maps.
    $n$ is the number of trials; $k$ is the number of successful runs.
    The best result in each environment is \textbf{bolded}. Success is defined as reaching within 5\,m of the goal without getting immobilized for greater than 10 seconds. 
  }
  \label{tab:success_rate}
  
  \begin{tabular}{l  cc  cc}
    \toprule
    & \multicolumn{2}{c}{\textbf{Hill (\textsc{Cola}\xspace)} ($n{=}20$)}
    & \multicolumn{2}{c}{\textbf{Ridge (\textsc{MerVan}\xspace)} ($n{=}50$)} \\
    \cmidrule(lr){2-3}\cmidrule(lr){4-5}
    \textbf{Method} & $k$ & SR\,(\%) & $k$ & SR\,(\%) \\
    \midrule
    \method      & \textbf{16} & \textbf{80} & \textbf{47} & \textbf{94} \\
    \textsc{dgt-map(Iso)}\xspace  & 10          & 50          & 45          & 90          \\
    \textsc{dgt-map(ST)}\xspace &  5          & 25          & 14          & 28          \\
    \baselineSlopeBased          & \textbf{16} & \textbf{80} & 26          & 52          \\
    \baselineBinary                & 10          & 50          & 20          & 40          \\
    \bottomrule
  \end{tabular}
\end{table}

\begin{table}[t]
  \renewcommand{\arraystretch}{1.35}
  \setlength{\tabcolsep}{5pt}
  \centering
  \caption{%
    \textbf{Navigation Success Rate on Random Start-and-Goal Tasks.}
    DGT-Map (multi-task) is compared against its single-task variant
    on a map that is similar to the training environment but was not seen during training.
    $n$ denotes the number of trials; $k$ the number of successes.
    The best result per vehicle is \textbf{bolded}. Success is defined as reaching within 5\,m of the goal without getting immobilized for greater than 10 seconds.
  }
  \label{tab:random_sr}
  \begin{tabular}{l  cc  cc}
    \toprule
    & \multicolumn{2}{c}{\textbf{\textsc{Jeep}\xspace} ($n{=}100$)}
    & \multicolumn{2}{c}{\textbf{\textsc{VwVan}\xspace} ($n{=}99$)} \\
    \cmidrule(lr){2-3}\cmidrule(lr){4-5}
    \textbf{Method} & $k$ & SR\,(\%) & $k$ & SR\,(\%) \\
    \midrule
    \method        & \textbf{70} & \textbf{70.0} & \textbf{54} & \textbf{54.5} \\
    \baselineSingleTask & 64          & 64.0          & 35          & 35.4          \\
    \bottomrule
  \end{tabular}
\end{table}

\begin{figure}
    \centering
    \includegraphics[width=1.0\linewidth]{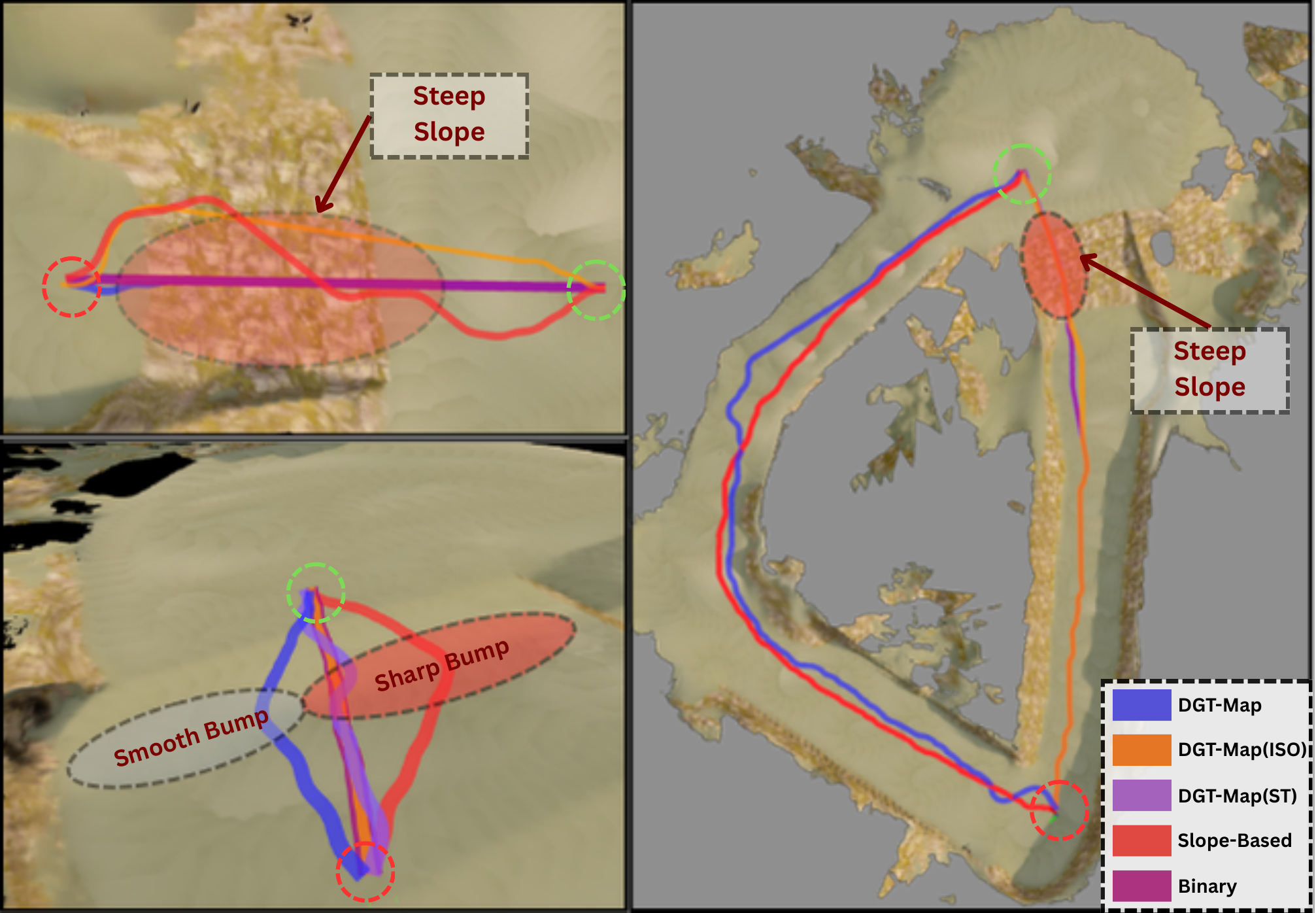}
    \caption{Planned paths in CARLA using 5 different costmaps with Hybrid A*.
    \textbf{Right:} Path planned for the \cola to ascend from base to summit.
    \method (blue) and \baselineSlopeBased (red) route around the steep slope via the longer path,
    correctly identifying that ascent through the steep slope is infeasible.
    The remaining methods attempt (and fail) to climb the steep slope directly.
    \textbf{Top left:} Paths planned to descend from the summit to the base.
    All methods, including \method, route through the steep section,
    confirming that \method distinguishes feasible descent from infeasible ascent on the same terrain.
    \textbf{Bottom left:} Ridge map scenario.
    \method routes through the smoothed region,
    while the other methods end up taking the less traversable path through the sharp bump.}
    \vspace{-1em}
    \label{fig:planned_paths}
\end{figure}

\textbf{Experiment 1:}
Both \method and \baselineSlopeBased achieve the highest success rates as seen in \autoref{tab:success_rate}. During ascent (\autoref{fig:planned_paths}, right), both methods route around the steep slope via the longer path, correctly identifying that the steep section is infeasible for uphill travel. The remaining methods attempt the steep section directly, leading to failures.

During descent (\autoref{fig:planned_paths}, top-left), all methods route through the steep section. This confirms that \method does not simply avoid steep terrain conservatively but correctly distinguishes feasible descent from infeasible ascent on the same terrain.

% The success rate of \baselineAverage and \baselineBinary suffer due to planning through the steep section on ascents . \baselineSingleTask performs worst, frequently selecting the infeasible uphill route and exhibiting additional noise in the costmap that produces unstable paths.

These results highlight the importance of preserving directional information when traversability varies with the direction of travel.

\textbf{Experiment 2:}
\method and \baselineAverage achieve the highest success rates in \autoref{tab:success_rate}. Because the sharp bump is less direction-dependent than the hill in Experiment~1, explicit heading conditioning provides less benefit in this scenario. \baselineBinary and \baselineSingleTask frequently fail to avoid the sharp bump, while \baselineSlopeBased models the approach slope but not the reduced traversability of the sharp edge. Accordingly, \method routes toward the smooth crossing in \autoref{fig:planned_paths} (bottom-left), whereas \baselineSlopeBased crosses the sharp bump. These results indicate that the learned representation captures terrain characteristics beyond simple geometric slope.

The vehicle-specific costmaps in \autoref{fig:sharp_hill_combined} further show that the \jeep assigns low cost to the sharp bump, while the \van assigns it substantially higher cost, consistent with their differing traversal capabilities. This demonstrates that \method preserves vehicle-specific terrain responses.

% \textbf{Experiment 2:}
% \method and \baselineAverage outperform other methods as seen in \autoref{tab:success_rate}. In this scenario, the sharp bump is less direction-dependent than the hill in Experiment 1, reducing the importance of explicit heading conditioning.

% \baselineBinary and \baselineSingleTask perform poorly, failing to avoid the sharp bump. \baselineSlopeBased achieves moderate performance, correctly modeling approach slopes but failing to detect the reduced traversability of the sharp edge. 

% This behavior is visible in \autoref{fig:planned_paths} (bottom-left), where \method routes toward the smooth bump crossing, while \baselineSlopeBased crosses at the sharp bump, and the remaining methods take the shortest path regardless of terrain difficulty.

% These results indicate that learned features capture terrain characteristics beyond simple geometric slope.

% Inspecting the costmaps generated by the \jeep (\autoref{fig:sharp_hill_combined}, bottom-left) and the \van (bottom-right), we see that the costmap for the \jeep ignores the bump while the costmap for the \van shows higher cost at the sharp bump as well as other sharp features in the map. This illustrates how \method makes vehicle specific predictions based on their differing abilities to traverse terrain features. 

\textbf{Experiment 3:}
\method training achieves higher observed success rates than \baselineSingleTask for both evaluated vehicles, increasing success from 64.0\% to 70.0\% for the \jeep and from 35.4\% to 54.5\% for the \vwVan (\autoref{tab:random_sr}). The larger absolute gain for the \vwVan is consistent with the hypothesis that the shared multi-task backbone provides a greater benefit when vehicle-specific training data is limited.

% Multi-task training outperforms \baselineSingleTask as seen in \autoref{tab:random_sr}. Two factors likely contribute to this improvement. First, shared training across vehicles regularizes the representation, encouraging the model to learn generalizable terrain features. Second, cross-vehicle transfer improves performance for less capable vehicles.

% In particular, the \vwVan benefits substantially from multi-task training. Because the \jeep collected more training samples and was able to traverse more challenging terrain during data collection, its data expands the range of terrain examples available during training. The resulting shared representation improves prediction quality for the \vwVan, suggesting effective transfer of terrain knowledge from a data-rich vehicle to a data-scarce vehicle.

% ============================================================================
% FIGURES
% ============================================================================

% \section{Conclusion}

% Off-road navigation requires traversability representations that reflect both the directional structure of terrain and the vehicle-dependent nature of traversal feasibility. 
% Existing scalar costmaps collapse anisotropic effects into a single value, while prior directional models have not demonstrated integration within kinodynamically feasible planning and closed-loop execution.

% In this work, we presented \method, a self-supervised framework for learning global, directional, vehicle-conditioned traversability maps from RGB-D observations. By predicting locomotion-derived costs for heading-aligned terrain patches and leveraging multi-task learning across heterogeneous vehicles, \method preserves both anisotropic terrain structure and platform-specific behavior within a unified representation.

% Through integration with a Hybrid A* planner and closed-loop control, we demonstrated consistent improvements in task success over geometric, binary, and learned isotropic baselines. These results show that explicitly modeling directionality and vehicle dependence at the map level leads to more reliable off-road navigation performance.

% We believe that directional, vehicle-conditioned traversability modeling provides a principled foundation for terrain-aware planning and offers a promising direction for scalable off-road autonomy.

\section{Conclusion}

In this work, we presented \method, a self-supervised framework for learning global, directional, and vehicle-conditioned traversability maps from RGB-D observations. By combining heading-aligned terrain representations, locomotion-derived supervision, and multi-task learning, \method captures traversal asymmetries and vehicle-specific terrain responses that cannot be represented by a single isotropic cost per location.

When integrated with Hybrid A* and closed-loop control, DGT-MAP improves navigation performance over the evaluated binary, slope-based, isotropic, and single-task baselines. The experiments show that DGT-MAP captures direction-dependent and vehicle-specific traversability, while multi-task training improves performance relative to single-task training for both evaluated vehicles. These results highlight the value of representing traversability as both directional and vehicle-dependent, enabling planners to distinguish not only where terrain is difficult, but also for which vehicle and from which direction.

\section{Limitations \& Future Work}

\textbf{Simulation-Only Training:}
\method is trained and evaluated entirely in simulation. 
Although CARLA provides controllable environments, its physics and terrain models do not fully capture real off-road phenomena such as deformable soil, loose rocks, or mud, and visual terrain diversity is limited. 
Future work will investigate sim-to-real transfer strategies, including domain randomization, fine-tuning with limited real-world data, and multi-task learning across simulated and real vehicles.

\textbf{Vehicle Scope:}
The current architecture uses vehicle-specific output heads and therefore assumes that the target vehicle is represented during training. Zero-shot generalization to unseen vehicle geometries or dynamics is not evaluated. Conditioning directly on vehicle properties is a direction for future work.

\textbf{Input-Modality Attribution:}
Although \method uses both RGB and height-derived features, the present experiments do not isolate the contribution of RGB. The ridge experiment demonstrates performance beyond the slope-based baseline, but not beyond geometry as a whole.

\textbf{Perfect Localization:} RGB-D data is combined with simulator ground-truth localization data to create the maps fed to the model. Real-world deployment would require integrating a localization framework, which would likely produce noisier results that could affect performance.

\textbf{Offline Costmap Generation:}
Directional costmaps are currently generated offline prior to planning, which limits applicability in previously unseen environments. 
Future work will optimize the implementation so it can be integrated into exploration-driven mapping systems that incrementally update traversability during navigation.

\textbf{Directional Inference Cost:}
Generating directional costmaps requires rotating the BEV input and running the CNN separately for each heading bin, increasing inference time linearly with the number of orientations. 
Equivariant neural networks~\cite{cohen2016gcnn,weiler2019e2cnn} that predict directional costs in a single forward pass may provide a more efficient solution.

\textbf{Lack of Terrain-Aware Control:}
The traversability costmap is used only during planning and is not incorporated into the low-level controller. 
As a result, terrain difficulty is not explicitly considered if the vehicle deviates from the planned trajectory. 
Future work will explore terrain-aware control strategies such as cost-augmented MPC.

% \input{sections/9.Aknowledgment}

%% References
%\bibliographystyle{IEEEtran_ShortURL}
\bibliographystyle{IEEEtran}
\bibliography{citations.bib}
\end{document}